\pdfoutput=1

\documentclass[11pt]{article}

\usepackage[]{ACL2023}

\usepackage{times}
\usepackage{latexsym}

\usepackage[T1]{fontenc}

\usepackage[utf8]{inputenc}

\usepackage{microtype}
\usepackage{algorithm}
\usepackage{algpseudocode}
\usepackage{booktabs}
\usepackage{amsmath}
\usepackage{multirow}
\usepackage{bbding}
\usepackage{mathrsfs}
\usepackage{subfigure}
\usepackage{amssymb}
\usepackage{mathtools}
\usepackage{graphicx} 

\definecolor{lightgreen}{HTML}{CEF6CE}
\definecolor{lightred}{HTML}{FD8F70}
\usepackage{url}

\usepackage{inconsolata}

\title{Decoupling Pseudo Label Disambiguation and Representation Learning for Generalized Intent Discovery}

\author{Yutao Mou$^{1*}$, Xiaoshuai Song$^{1*}$, Keqing He$^{2*}$,Chen Zeng$^{1}$,Pei Wang$^{1}$ \\
{\bf Jingang Wang$^{2}$,} {\bf Yunsen Xian$^{2}$,} {\bf Weiran Xu$^{1}$}\thanks{\ \ The first three authors contribute equally. Weiran Xu is the corresponding author.}\\
  $^1$Beijing University of Posts and Telecommunications, Beijing, China\\
$^{2}$Meituan, Beijing, China\\
  \texttt{\{myt,songxiaoshuai,chenzeng,wangpei,xuweiran\}@bupt.edu.cn}\\
  \texttt{\{hekeqing,wangjingang,xianyunsen\}@meituan.com}
}
\begin{document}
\maketitle
\begin{abstract}
Generalized intent discovery aims to extend a closed-set in-domain intent classifier to an open-world intent set including in-domain and out-of-domain intents. The key challenges lie in pseudo label disambiguation and representation learning. Previous methods suffer from a coupling of pseudo label disambiguation and representation learning, that is, the reliability of pseudo labels relies on representation learning, and representation learning is restricted by pseudo labels in turn. In this paper, we propose a decoupled prototype learning framework (DPL) to decouple pseudo label disambiguation and representation learning. Specifically, we firstly introduce prototypical contrastive representation learning (PCL) to get discriminative representations. And then we adopt a prototype-based label disambiguation method (PLD) to obtain pseudo labels. We theoretically prove that PCL and PLD work in a collaborative fashion and facilitate pseudo label disambiguation. Experiments and analysis on three benchmark datasets show the effectiveness of our method.\footnote{We  release our code at \url{https://github.com/songxiaoshuai/DPL}}


\end{abstract}

\section{Introduction}


Intent classification (IC) is an important component of task-oriented dialogue (TOD) systems. Traditional intent classification models are based on a closed-set hypothesis \cite{chen2019bert, yang2021generalized}. That is, they rely on a pre-defined intent set provided by domain experts and can only recognize limited in-domain (IND) intent categories. However, users may input out-of-domain (OOD) queries in the real open world. OOD intent detection \cite{lin-xu-2019-deep, xu-etal-2020-deep, zeng-etal-2021-modeling, wu-etal-2022-revisit, Wu2022DisentanglingCS} identifies whether a user query falls outside the range of pre-defined IND intent set. Further, OOD intent discovery task \cite{Lin2020DiscoveringNI, Zhang2022NewID, mou-etal-2022-disentangled, mou2022watch} (also known as new intent discovery) groups unlabeled OOD intents into different clusters. However, all these work cannot expand the recognition scope of the existing IND intent classifier incrementally.

\begin{figure}[t]
    \centering
    \resizebox{.48\textwidth}{!}{
    \includegraphics{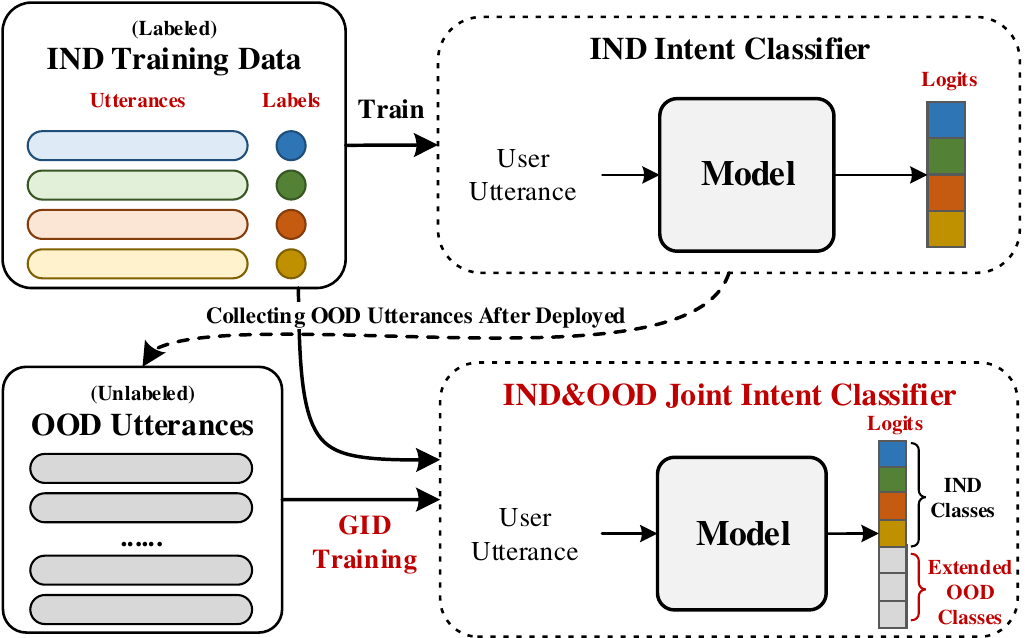}}
    \caption{The illustration of GID task.}
    \label{fig:GID task}
    \vspace{-0.65cm}
\end{figure}

To solve the problem, \citet{mou-etal-2022-generalized} proposes the Generalized Intent Discovery (GID) task, which aims to simultaneously classify a set of labeled IND intents while discovering and recognizing new unlabeled OOD types incrementally. As shown in Fig \ref{fig:GID task}, GID extends a closed-set IND classifier to an open-world intent set including IND and OOD intents and enables the dialogue system to continuously learn from the open world. Previous GID methods can be generally classified into two types: pipeline and end-to-end. The former firstly performs intent clustering and obtains pseudo OOD labels using K-means \cite{MacQueen1967SomeMF} or DeepAligned \cite{Zhang2021DiscoveringNI}, and then mixes labeled IND data with pseudo-labeled OOD data to jointly learn a new classifier. However, pipeline-based methods separate the intent clustering stage from the joint classification stage and these pseudo OOD labels obtained in the intent clustering stage may induce severe noise to the joint classification. In addition, the deep semantic interaction between the labeled IND intents and the unlabeled OOD data is not fully considered in the intent clustering stage. To alleviate these problems, \citet{mou-etal-2022-generalized} proposes an end-to-end (E2E) framework. It mixes labeled IND data with unlabeled OOD data in the training process and simultaneously learns pseudo OOD cluster assignments and classifies IND\&OOD classes via self-labeling \cite{asano2020self}.




\begin{figure}[t]
    \centering
    \resizebox{.5\textwidth}{!}{
    \includegraphics{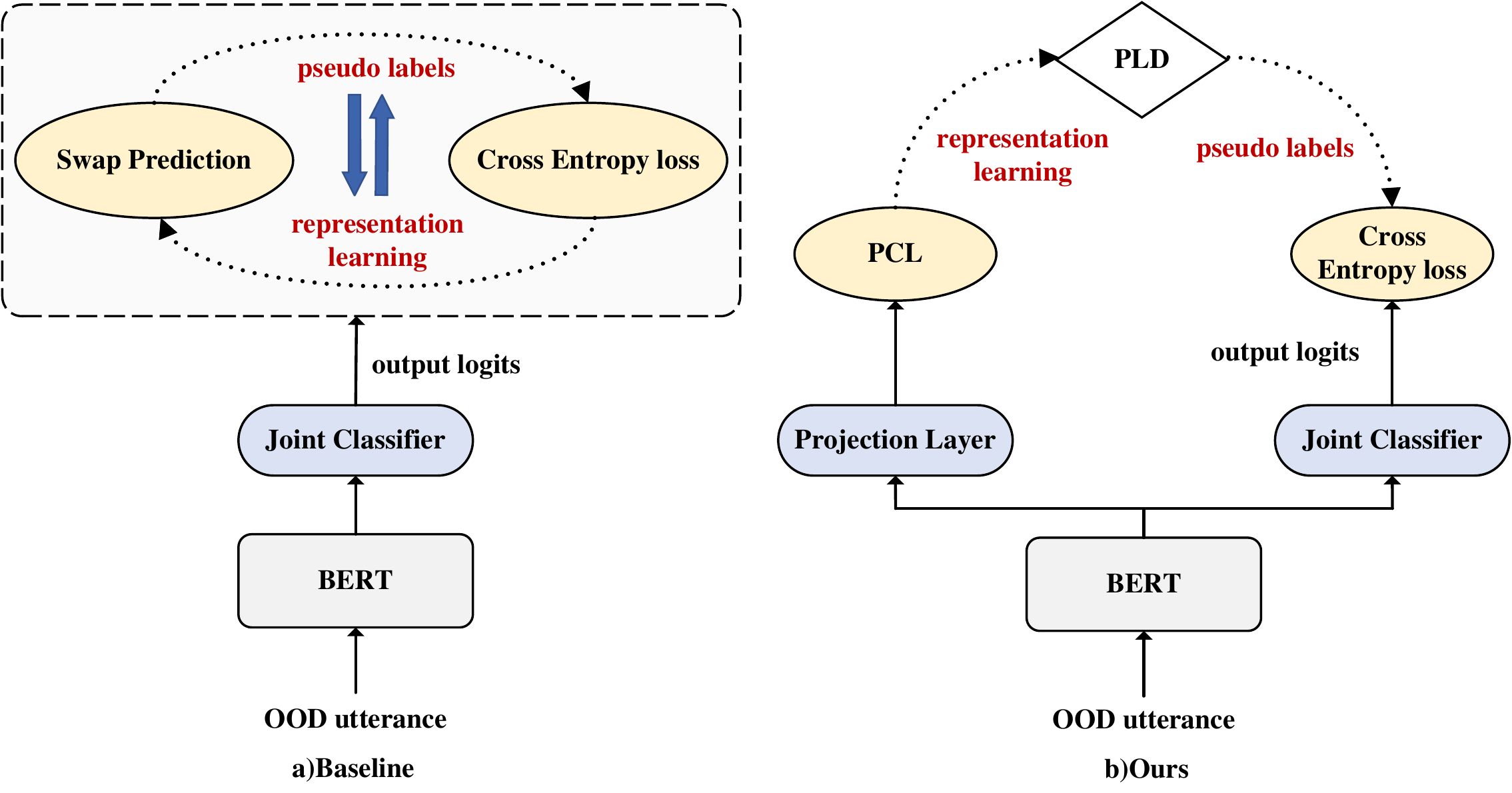}}
    \vspace{-0.7cm}
    \caption{Comparison between baseline E2E and our proposed DPL method. }
    \label{fig:motivation}
    \vspace{-0.65cm}
\end{figure}

E2E framework achieves state-of-the-art results in most scenarios, but there are still two key challenges: (1) \textbf{Pseudo Label Disambiguation}. In the GID task, the performance of the joint classifier depends on pseudo labels of unlabeled OOD data, so we need to improve the reliability of pseudo labels during the training process, which is called "pseudo label disambiguation".
(2) \textbf{Representation Learning}. We hope to form a clear cluster boundary for different IND and OOD intent types, which also benefits pseudo label disambiguation.
As shown in Fig \ref{fig:motivation}(a), the state-of-the-art E2E method \cite{mou-etal-2022-generalized} adopts a self-labeling strategy \cite{asano2020self, Fini2021AUO} for pseudo label disambiguation and representation learning. Firstly, it obtains the pseudo label of an OOD query by its augmented view in a swapped prediction way for pseudo label disambiguation. Next, it uses the pseudo labels as supervised signals and adopts a cross-entropy classification loss for representation learning.
Therefore, pseudo label disambiguation and representation learning are coupled, which has led to a non-trivial dilemma: the inaccurate pseudo labels will limit the quality of representation learning, and poor representation quality will in turn prevent effective pseudo label disambiguation. We also find that the coupling of pseudo label disambiguation and representation learning leads to slow convergence of the model (see Section \ref{Pseudo label}).

To solve this problem, we propose a novel \textbf{D}ecoupled \textbf{P}rototype \textbf{L}earning framework (\textbf{DPL}) for generalized intent discovery, which aims to decouple pseudo label disambiguation and representation learning. 
Different from the previous E2E method, DPL consists of two complementary components: prototypical contrastive representation learning (PCL) to get good intent representations and prototype-based label disambiguation (PLD) to obtain high-quality pseudo labels, as shown in Fig \ref{fig:motivation}(b). In our framework, PCL and PLD work together to realize the decoupling of pseudo label disambiguation and representation learning. Specifically, we firstly employ the output probability distribution of the joint classifier to align samples and corresponding prototypes and perform prototypical contrastive representation learning \cite{PCL, wang-etal-2021-bridge, cui-etal-2022-prototypical}. We aim to pull together similar samples to the same prototype and obtain discriminative intent representations. Secondly, based on the embeddings and class prototypes learned by PCL, we introduce a prototype-based label disambiguation, which gradually updates pseudo labels based on the class prototypes closest to the samples. Finally, we use these pseudo labels to train a joint classifier.
We leave the details in the following Section \ref{approach}. In addition, we theoretically explain that prototypical contrastive representation learning gets closely aligned representations for examples from the same classes and facilitates pseudo label disambiguation (Section \ref{theory}). We also perform exhaustive experiments and qualitative analysis to demonstrate that our DPL framework can obtain more reliable pseudo labels and learn better representations in Section \ref{qualitative}.



Our contributions are three-fold: (1) We propose a novel decoupled prototype learning (DPL) framework for generalized intent discovery to better decouple pseudo label disambiguation and representation learning. (2) We give a theoretical interpretation of prototypical contrastive representation learning to show that it gets better representations to help pseudo label disambiguation.
(3) Experiments and analysis on three benchmark datasets demonstrate the effectiveness of our method for generalized intent discovery. 





\section{Approach}
\label{approach}

\subsection{Problem Formulation}
\label{definition}
Given a set of labeled in-domain data $\textbf{D}^{IND}=\left\{\left({x}_{i}^{IND}, {y}_{i}^{IND}\right)\right\}_{i=1}^n$ and unlabeled OOD data $\textbf{D}^{OOD}=\left\{\left({x}_{i}^{OOD}\right)\right\}_{i=1}^m$, where ${y}_{i}^{IND} \in \mathcal{Y}^{IND}, \mathcal{Y}^{IND}=\{1,2, \ldots, N\}$, GID aims to train a joint classifier to classify an input query to the total label set $\mathcal{Y}=\left\{1, \ldots, N, N+1, \ldots, N+M\right\}$ where the first $N$ elements denote labeled IND classes and the subsequent $M$ ones denote newly discovered unlabeled OOD classes. For simplicity, we assume the number of OOD classes is specified as $M$. Since OOD training data is unlabeled, how to obtain accurate pseudo labels a key problem.


\subsection{Overall Architecture}
\begin{figure}[t]
    \centering
    \resizebox{.5\textwidth}{!}{
    \includegraphics{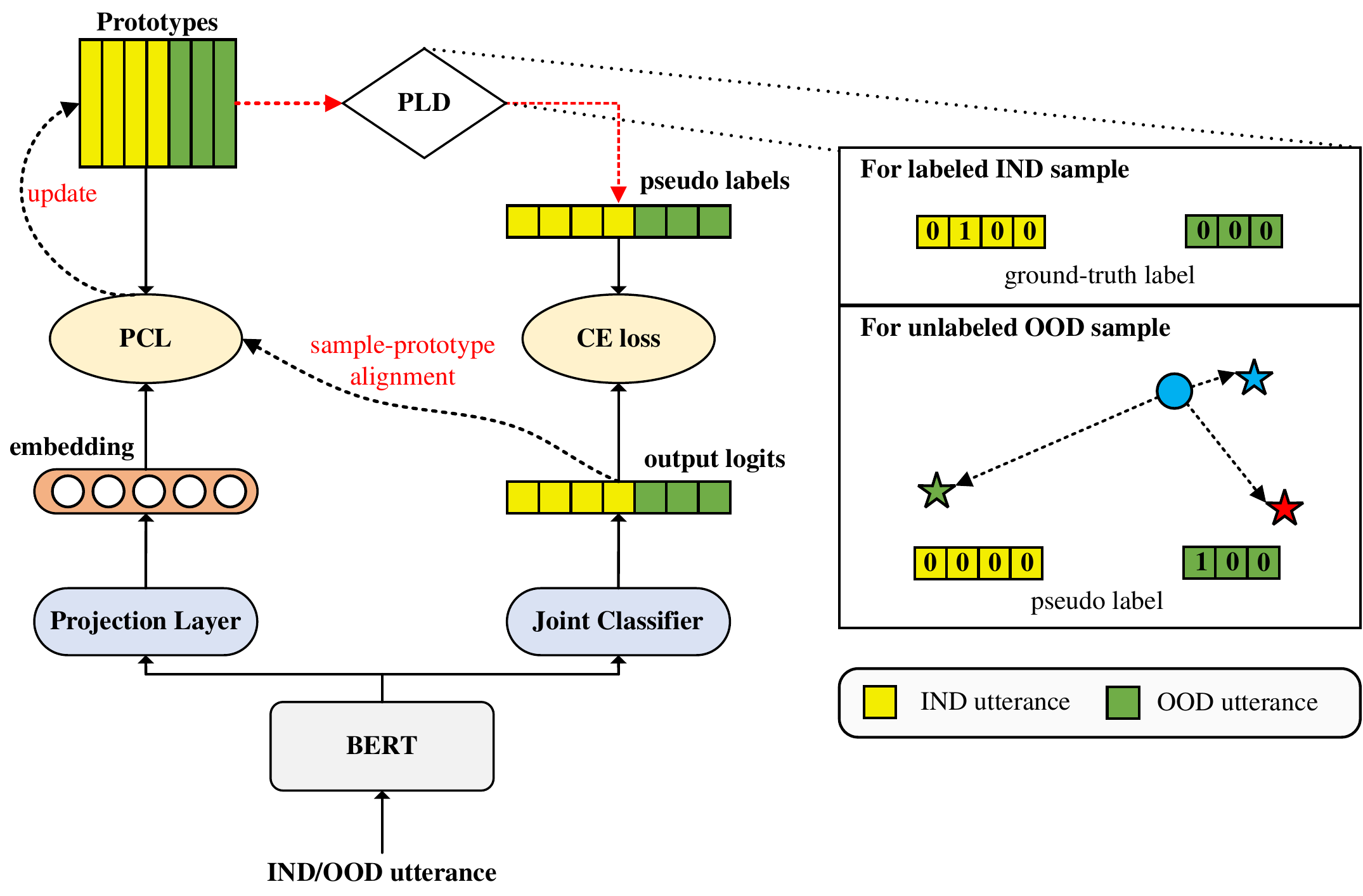}}
    \vspace{-0.75cm}
    \caption{Overall architecture of our DPL method.}
    \label{fig:model}
    \vspace{-0.6cm}
\end{figure}

Fig \ref{fig:model} displays the overall architecture of our proposed decoupled prototype learning (DPL) framework for generalized intent discovery. We firstly get contextual features using BERT encoder \cite{devlin-etal-2019-bert}. To better leverage prior intent knowledge, we first pre-train the encoder on labeled IND data to get intent representations as E2E \cite{mou-etal-2022-generalized}. And then we add a joint classifer and a projection layer \footnote{In the experiments, we use a two-layer non-linear MLP to implement the projection layer.} on top of BERT. Given an input query, the projection layer maps the intent features of BERT encoder to a hypersphere, and uses prototypical contrastive representation learning (PCL) to further learn discriminative intent representations and class prototypes. Based on the representations and prototypes, we adopt a prototype-based label disambiguation (PLD) method to obtain pseudo labels, and use a cross-entropy(CE) objective to optimize the joint classifier.
In the DPL framework, prototypical contrastive representation learning is not limitted by pseudo labels, and decouples pseudo label disambiguation and representation learning. We provide a pseudo-code of DPL in Appendix \ref{appendix_algorithm}.


\subsection{Prototypical Contrastive Learning}

\textbf{Sample-prototype alignment} We introduce prototypical contrastive representation learning (PCL) in our DPL framework. Firstly, we randomly initialize the $L_{2}$-normalized prototype embedding $\mu_{j}, j = 1,2,...,N+M$ of each intent category, which can be seen as a set of representative embedding vectors, and then for each input sample $x_{i}$, we need to align it with the corresponding class prototype. Specifically, if $x_{i}$ belongs to IND intents, we use ground-truth label to align the sample with class prototype. If the input sample belongs to OOD intents, the output logit $l_{i}^{OOD}=(l_{i}^{N+1},...,l_{i}^{N+M})$ can be obtained by the joint classifier $\mathit{f}(x_{i})$ \footnote{Following \cite{mou-etal-2022-generalized}, we also adopt SK algorithm \cite{Cuturi2013SinkhornDL} to calibrate the output logits.}, and we can use $l_{i}^{OOD}$ to  align the sample with class prototype.
The alignment relationship is as follows:
{
\setlength{\abovedisplayskip}{0.1cm}
\setlength{\belowdisplayskip}{0.1cm}
\begin{equation}
\label{alignment}
\boldsymbol{q_i}= \begin{cases}{\left[y_{i}^{IND}; \mathbf{0}_{M}\right]} & {x_i} \in \textbf{D}^{IND} \\ {\left[\mathbf{0}_{N}; l_{i}^{OOD}\right]} & {x_i} \in \textbf{D}^{OOD}\end{cases}
\end{equation}

where $y_{i}^{IND}$ is a one-hot vector of ground-truth label, $\mathbf{0}_{M}, \mathbf{0}_{N}$ are M or N-dimention zero vectors and $q_{i}^{j}, j=1,2,...,N+M$ represents the confidence probability that sample $x_{i}$ belongs to prototype $\mu_{j}$.
After obtaining the alignment relationship between samples and prototypes, we get the $L_{2}$-normalized embedding $z_{i}$ of sample $x_{i}$ through the projection layer $\mathit{g}(x_{i})$, and then perform prototypical contrastive learning as follows:
\begin{equation}
\mathcal{L}_{P C L}=-{\sum_{i, j} q_{i}^{j} \log \frac{\exp \left(\operatorname{sim}\left({z}_{i}, \mu_{j}\right) / \tau\right)}{\sum_{r} \exp \left(\operatorname{sim}\left({z}_{i}, \mu_{r}\right) / \tau\right)}}
\end{equation}


where $\tau$ denotes temperature, and we set it to 0.5 in our experiments. PCL pulls together similar samples to the same prototype and obtain discriminative intent representations. Furthermore, we also add the instance-level contrastive loss to alleviate the problem of incorrect alignment between samples and prototypes caused by unreliable confidence probability at the beginning of training. 
\begin{equation}
\mathcal{L}_{ins}=-{\sum_{i} \log \frac{\exp \left(\operatorname{sim}\left({z}_{i}, \hat{{z}}_{i}\right) / \tau\right)}{\sum_{k} \mathbf{1}_{[k \neq i]} \exp \left(\operatorname{sim}\left({z}_{i}, {z}_{k}\right) / \tau\right)}}
\end{equation}

where $\hat{{z}}_{i}$ denotes the dropout-augmented view of ${z}_{i}$. Finally, we jointly optimize $\mathcal{L}_{PCL}$ and $\mathcal{L}_{ins}$ to learn cluster-friendly representation.

\textbf{Update prototype embedding} The class prototype embedding needs to be constantly updated during the training process. The naive way to update prototypes is to calculate the average value of embeddings for samples of the same class at each iteration. However, this will lead to a large amount of computing overhead, which will lead to unbearable training delays. Therefore, we update the prototype vector in a moving-average style:
\begin{equation}
\label{update}
\begin{aligned}
\boldsymbol{\mu}_c = \operatorname{Normalize}\left(\gamma \boldsymbol{\mu}_c+(1-\gamma) \boldsymbol{{z}_{i}}\right)
\end{aligned}
\end{equation}
where the prototype ${\mu}_c$ of intent class $c$ can be defined as the moving-average of normalized embeddings ${z}_{i}$, if the confidence of sample $x_{i}$ belonging to category $c$ is the largest. The moving average coefficient $\gamma$ is a tunable hyperparameter.


\subsection{Prototype-based Label Disambiguation}
Prototypical contrastive learning gets discriminative intent representations, compact cluster distributions and class prototype embeddings that fall in the center of corresponding clusters. Next, we need to use the learned class prototypes for pseudo label disambiguation. Specifically, if an input sample $x_{i}$ belongs to IND intents, we use ground-truth label directly, if an input sample belongs to OOD intents, the pseudo target assignment is to find the nearest prototype of the current embedding vector. The pseudo label is constructed as follows:
{
\setlength{\abovedisplayskip}{0.1cm}
\setlength{\belowdisplayskip}{0.1cm}
\begin{equation}
\label{disambiguation_1}
\boldsymbol{y}_{i}= \begin{cases}{\left[{y}_{i}^{IND}; \mathbf{0}_{M}\right]} & {x}_{i} \in \textbf{D}^{IND} \\ {\left[\mathbf{0}_{N}; \hat{\boldsymbol{p}}_{i}^{OOD}\right]} & {x}_{i} \in \textbf{D}^{OOD}\end{cases} 
\end{equation}

\begin{equation}
\label{disambiguation_2}
\hat{\boldsymbol{p}}_{i}^{c}= \begin{cases}1 & \text { if } c=\arg \max _{j \in \mathcal{Y}^{OOD}} \boldsymbol{{z}_{i}}^{\top} \boldsymbol{\mu}_j \\ 0 & \text { else }\end{cases}
\end{equation}

After obtaining pseudo labels, we use cross-entropy loss $\mathcal{L}_{CE}$ to optimize the joint classifier, and learn to classify labeled IND intents and the newly discovered unlabeled OOD intents.


\section{Theoretical Analysis}
\label{theory}
In this section, we provide a theoretical explanation of why prototypical contrastive representation learning can learn cluster-friendly intent representations and class prototypes that facilitate pseudo label disambiguation. PCL essentially draws similar samples towards the same prototype, and forms compact clusters in the representation space, which is consistent with the goal of clustering, so we will explain it from the perspective of EM algorithm.

As defined in Section \ref{definition}, we have $n$ labeled IND samples and $m$ unlabeled OOD samples. In the GID task, our goal is to find suitable network parameters to maximize the log-likelihood function as follows:
\begin{equation}
\theta^*=\arg \max _\theta \sum_{i=1}^{n+m} \log P\left(x_i \mid \theta\right)
\end{equation}

\textbf{E-step} In the supervised learning setting, it is easy to estimate the likelihood probability using ground-truth labels. However, in the GID task, we not only have labeled IND samples, but also have a large number of unlabeled OOD samples, so we need to associate each sample with an implicit variable ${j}, j=1,2,...,N+M$ ($j$ represents the intent category). In addition, this likelihood function is hard to be directly optimized, so we need to introduce a probability density function ${q}_{i}(j)$ to represent the probability that sample ${x}_{i}$ belongs to intent category $j$. Finally, we can use Jensen’s inequality to derive the lower bound of the maximum likelihood function as follows (We leave detailed derivation process in appendix \ref{appendix_theory}):
\begin{equation}
\label{e-step}
\begin{aligned}
\theta^* & =\underset{\theta}{\operatorname{argmax}} \sum_{i=1}^{n+m} \log P\left(x_i \mid \theta\right) \\
& \geqslant \underset{\theta}{\arg \max } \sum_{i=1}^{n+m} \sum_{j \in y_{a l l}} q_i(j) \log \frac{P\left(x_i, j \mid \theta\right)} {q_{i}(j)} \\
\end{aligned}
\end{equation}

Since $\log (\cdot)$ is a concave function, the inequality holds with equality when $\frac{P\left(x_i, j \mid \theta\right)}{q_{i}(j)}$ is constant. Thus we can derive ${q}_{i}(j)$ as follows:
\begin{equation}
q_i(j)=\frac{P\left(x_i, j \mid \theta\right)}{\sum_{j \in y_{a l l}} P\left(x_i, j \mid \theta\right)}=P\left(j \mid x_i, \theta\right)
\end{equation}


We can know that when $q_i(j)$ is a posterior class probability, maximizing the lower bound of the likelihood function is equivalent to maximizing the likelihood function itself.
In our GID task, there are both labeled IND data and unlabeled OOD data. Therefore, for labeled IND data, we can directly use ground-truth label to estimate the posterior class probability. For unlabeled OOD data, we can estimate the posterior probability distribution by the joint classifier.
This provides theoretical support for the sample-prototype alignment in PCL.


\textbf{M-step} We have estimated $q_i(j)$ in E-step. Next, we need to maximize the likelihood function and find the optimal network parameters under the assumption that $q_i(j)$ is known. The optimization objective is as follows (We leave detailed derivation process in appendix \ref{appendix_theory}):
\begin{equation}
\label{m-step}
\begin{aligned}
 & L(\theta) =\max \sum_{i=1}^{n+m} \sum_{j \in y_{a l l}} q_i(j) \log \frac{P\left(x_i, j \mid \theta\right)}{q_i(j)} \\
& \approx \max \sum_{i=1}^{n+m} \sum_{j \in y_{a l l}} q_i(j) \log P\left(x_i \mid j, \theta\right) \\
& \approx  \max \sum_{i=1}^{n+m} \sum_{j \in y_{a l l}} q_{i(j)} \log \frac{\exp \left(\frac{\boldsymbol{z_i} \cdot \boldsymbol{\mu_{j}}}{\sigma_j^2}\right)}{\sum_{r \in y_{a l l}} \exp \left(\frac{\boldsymbol{z_i} \cdot \boldsymbol{\mu_{r}}}{\sigma_r^2}\right)} \\
& \Leftrightarrow \min \mathcal{L}_{P C L} \\
\end{aligned}
\end{equation}

\begin{table*}[t]
\centering
\resizebox{0.95\textwidth}{!}{%
\begin{tabular}{l||c|cc|cc||c|cc|cc||c|cc|cc}
\hline
\multirow{3}{*}{Method} & \multicolumn{5}{c||}{GID-SD} & \multicolumn{5}{c||}{GID-CD} & \multicolumn{5}{c}{GID-MD} \\ \cline{2-16}
                        & IND                  & \multicolumn{2}{c|}{OOD} & \multicolumn{2}{c||}{ALL} & IND                  & \multicolumn{2}{c|}{OOD} & \multicolumn{2}{c||}{ALL} & IND                  & \multicolumn{2}{c|}{OOD} & \multicolumn{2}{c}{ALL} \\
                        & ACC                  & ACC         & F1        & ACC         & F1        & ACC                  & ACC         & F1        & ACC         & F1        & ACC                  & ACC         & F1        & ACC         & F1        \\ \hline 
 k-means                          & 90.38    & 62.34     & 62.44    & 78.99 & 78.32   &97.70	&61.67	&60.43	&83.20	&82.30 & 97.26     & 73.00    & 72.66    & 87.56    & 87.08  \\ \hline
 DeepAligned  &91.72	&69.11	&69.72	&82.57	&82.10 &97.85	&78.55	&77.81	&90.12	&89.68  &97.85	&87.55	&87.14	&93.70	&93.29                   \\ \hline
  DeepAligned-Mix 	&82.30	&54.97	&59.79	&71.30	&69.60 &97.33	&72.41	&71.54	&87.36	&86.21	&92.86	&81.70	&83.30	&88.12	&87.42                             \\ \hline
  End-to-End  &92.84	&72.28	&73.28	&84.49	&84.10 &98.00	&79.19	&79.06	&90.46	&90.28  &98.32	&91.92	&92.46	&95.78	&95.73                  \\ \hline
  DPL(ours)   &92.89	&\textbf{74.38}	&\textbf{75.46} &\textbf{85.43} &\textbf{85.34} &98.37	&\textbf{82.40}	&\textbf{82.37} &\textbf{91.98} &\textbf{91.85} &98.29 &\textbf{92.84} &\textbf{93.00} &\textbf{96.11} &\textbf{95.96}    \\ \hline
\end{tabular}
}
\caption{Performance comparison on three benchmark datasets. Results are averaged over three random run.(p < 0.01 under t-test). Here we report the results of 40\% OOD ratio. For experimental results of more OOD ratios, we have made further discussion in Section \ref{OOD ratio}.}
\label{tab:main_result}
\end{table*}

where $P\left(x_i \mid j, \theta\right)$ represents the data distribution of the class $j$ in the representation space. We think that the larger the likelihood probability, the more reliable the pseudo labels.
We assume that the class $j$ follows a gaussian distribution in the representation space, and can derive that minimizing the PCL objective is equivalent to maximizing the likelihood function, which explains why the prototypical contrastive representation learning facilitates pseudo label disambiguation.

\section{Experiments}
\subsection{Datasets}
We conducted experiments on three benchmark datasets constructed by \cite{mou-etal-2022-generalized}, GID-SD(single domain), GID-CD(cross domain) and GID-MD(multiple domain).  GID-SD randomly selects intents as the OOD type from the single-domain dataset Banking \cite{casanueva-etal-2020-efficient}, which contains 77 intents in banking domain, and the rest as the IND type. GID-CD restricts IND and OOD intents from non-overlapping domains from the multi-domain dataset CLINC \cite{larson-etal-2019-evaluation}, which covers 150 intents in 10 domains, while GID-MD ignores domain constraints and randomizes all CLINC classes into IND sets and OOD sets. To avoid randomness, we average the results in three random runs. We leave the detailed statistical information of datasets to Appendix \ref{appendix_data}.

\subsection{Baselines}
Similar with \cite{mou-etal-2022-generalized}, we extensively compare our method with the following GID baselines: k-means \cite{MacQueen1967SomeMF}, DeepAligned \cite{Zhang2021DiscoveringNI}, DeepAligned-Mix \cite{mou-etal-2022-generalized}, End-to-End (E2E) \cite{mou-etal-2022-generalized}, in which E2E is the current state-of-the-art method for GID task. For a fair comparison, all baselines use the same BERT encoder as the backbone network. We leave the details of the baselines in Appendix \ref{appendix_baseline}. We adopt two widely used metrics to evaluate the performance of the joint classifier: Accuracy(ACC) and F1-score(F1), in which ACC is calculated over IND, OOD and total(ALL) classes respectively and F1 is calculated over OOD and all classes to better evaluate the ability of methods to discover and incrementally extend OOD intents. OOD and ALL ACC/F1 are the main metrics.

\subsection{Implementation Details}
\label{appendix_implementation}
For a fair comparison of the various methods, we use the pre-trained BERT model (bert-base-uncased \footnote{https://github.com/google-research/bert},  with 12-layer transformer) as our network backbone, and add a pooling layer to get intent representation(dimension=768). Moreover, we freeze all but the last transformer layer parameters to achieve better performance with BERT backbone and speed up the training procedure as suggested in \cite{Zhang2021DiscoveringNI}.

The class prototype embedding(dimension=128) is obtained by the representation through a linear projection layer.
For training, we use SGD with momentum as the optimizer, with linear warm-up and cosine annealing (  $lr_{min}$ = 0.01; for GID-SD, $lr_{base}$ = 0.02, for GID-CD and GID-MD,$lr_{base}$ = 0.1), and weight decay is 1.5e-4.The moving average coefficient $\gamma$=0.9.We train 100 epochs and use the Silhouette Coefficient(SC) of OOD data in the validation set to select the best checkpoints. Notably, We use dropout to construct augmented examples and the dropout value is fixed at 0.1. 

The average value of the trainable model parameters is 9.1M and the total parameters are 110M which is basically the same as E2E. In the training stage, the decoupling-related components of DPL bring approximately 8\% additional training load compared to E2E. In the inference stage, DPL only requires the classifier branch, without additional computational overhead. It can be seen that our DPL method has significantly improved performance compared to E2E, but the cost of time and space complexity is not large. All experiments use a single Nvidia RTX 3090 GPU(24 GB of memory).

\subsection{Main Results}
\label{main results}

Table \ref{tab:main_result} shows the performance comparison of different methods on three benchmark GID datasets. In general, our DPL method consistently outperforms all the previous baselines with a large margin in various scenarios. Next, we analyze the results from three aspects:

(1) \textbf{Comparison of different methods.} We can see that our proposed DPL method is better than all baselines. For example, DPL is superior to E2E by 2.1\% (OOD ACC), 2.18\% (OOD F1) and 1.24\% (ALL F1) on GID-SD dataset, 3.21\% (OOD ACC), 3.31\% (OOD F1) and 1.57\% (ALL F1) on GID-CD dataset, 0.92\% (OOD ACC), 0.54\% (OOD F1) and 0.23\% (ALL F1) on GID-MD dataset. This shows that DPL framework decouples pseudo label disambiguation and representation learning, which makes the pseudo labels and representation learning no longer restrict each other, and effectively improves the reliability of pseudo labels (We give a detailed analysis in section \ref{Pseudo label}). Accurate pseudo labels further improve the classification performance of the joint classifier, especially the ability to discover and recognize new OOD intent categories.

(2) \textbf{Single-domain scenario} Since IND and OOD intents belong to the same domain in GID-SD, and the difference between intents is smaller than that of multiple-domain dataset GID-MD, so it is more difficult to form clear cluster boundaries, and the performance of joint classifier is relatively low. Interestingly, we observed that the improvement of DPL method in single-domain dataset GID-SD is more significant than that in multiple-domain dataset GID-MD. For example, In GID-MD, DPL only increased by 0.54\% (OOD F1) compared with E2E, while in the more challenging GID-SD dataset, it increased by 2.18\% (OOD F1).
We believe that it is because prototypical contrastive representation learning can draw similar samples to the same prototype to learn cluster-friendly representation, which helps to form a clearer cluster boundary for each intents and improve the accuracy of pseudo labels. We leave more detailed analysis in Section \ref{representation}.

(3) \textbf{Cross-domain scenario} Since IND and OOD intents come from different domains in GID-CD, which means that it is more difficult to transfer the prior knowledge of labeled IND intents to help pseudo labeling of unlabeled OOD data. This can be seen from the small improvement (0.64\% OOD ACC) of E2E compared with DeepAligned. However, we find that our DPL method increased by 3.31\% (OOD F1) and 1.57\% (ALL F1) on GID-CD, which is far higher than the previous improvement.
We believe that this may be due to the use of prototypical contrastive representation learning to learn the class prototypes of IND and OOD intents at the same time, which more effectively make use of the prior knowledge of labeled IND intents to help the representation learning and obtain more accurate pseudo labels.

\section{Qualitative Analysis}
\label{qualitative}

\subsection{Pseudo Label Disambiguation}
\label{Pseudo label}

\begin{figure}[t]
    \centering
    \resizebox{.40\textwidth}{!}{
    \includegraphics{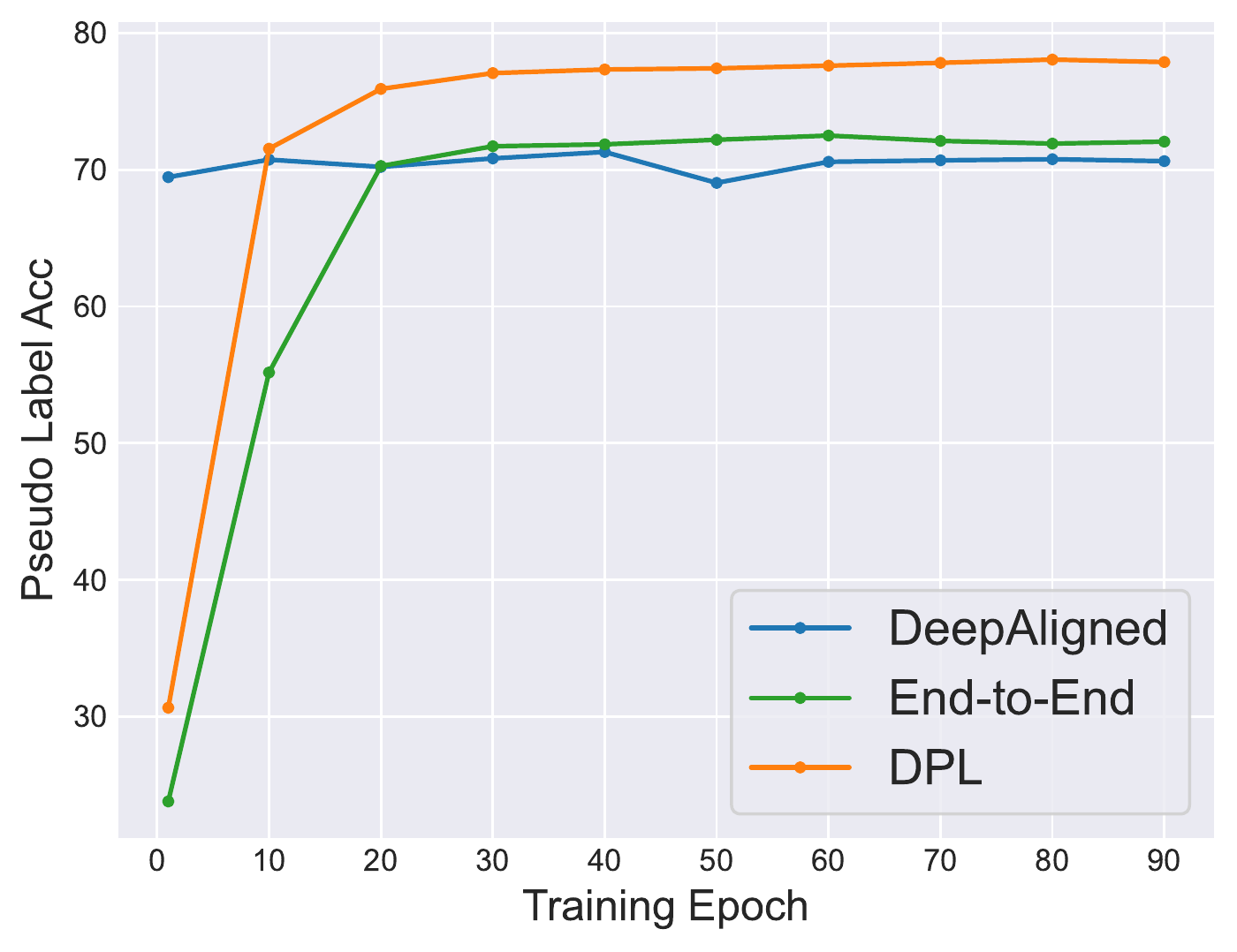}}
    \caption{Pseudo label accuracy curves in the training process.}
    \label{pseudo_label_acc}
\end{figure}

One of the key challenges of generalized intent discovery is pseudo label disambiguation. We compared the pseudo labels accuracy of different methods, as shown in Fig \ref{pseudo_label_acc}. Firstly, we can see that the end-to-end framework (DPL and E2E) has a higher upper bound of pseudo label accuracy than the pipeline framework (DeepAligned). We think that it is because the end-to-end framework fully considers the knowledge interaction between labeled IND intents and unlabeled OOD data in the training process. Next, we analyze the advantages of DPL over E2E in pseudo label disambiguation from two perspectives: (1) Our DPL method converges faster than E2E method. We think that the E2E method converges slower because pseudo label disambiguation and representation learning are coupled. Inaccurate pseudo labels limit the representation learning, while poor intent representation hinders pseudo label disambiguation. In contrast, our DPL method decouples the pseudo-label disambiguation and representation learning, which makes the pseudo labels and intent representation no longer restrict each other and accelerates the convergence.
(2) Compared with E2E method, our DPL method can obtain more accurate pseudo labels in the training process, and reach higher upper bound. We believe that there are two reasons for this. First, DPL framework decouples pseudo label disambiguation and representation learning. The quality of pseudo labels will not limit representation learning, so it can obtain more discriminative representation, thus improving the accuracy of pseudo labels. Besides, we use prototype-based contrastive learning for representation learning, which aligns with the subsequent prototype-based label disambiguation.



\subsection{Representation Learning}
\label{representation}

\begin{figure}[t]
    \centering
    \resizebox{.50\textwidth}{!}{
    \includegraphics{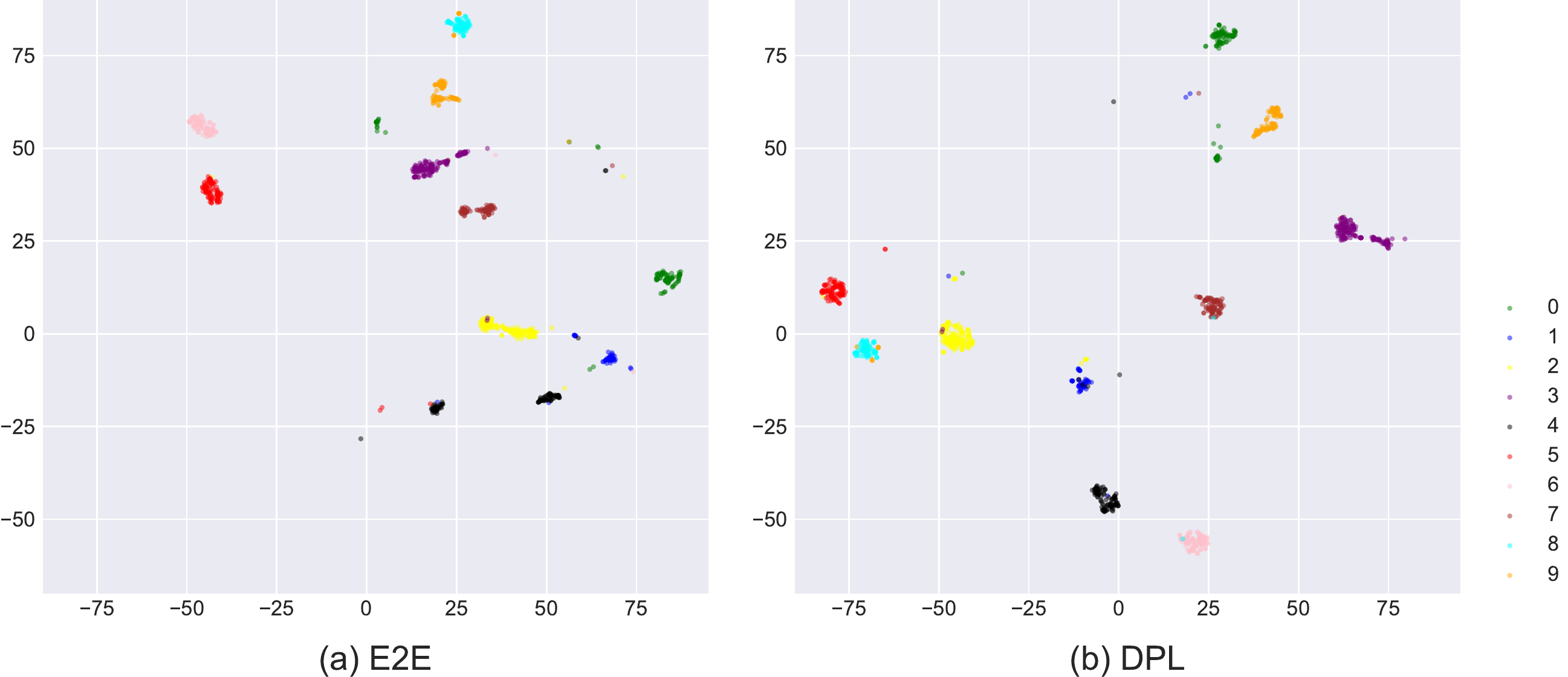}}
    \caption{Visualization of different methods.}
    \label{final_visualization}
\end{figure}

\begin{table}[t]
\centering
\resizebox{0.46\textwidth}{!}{%
\begin{tabular}{l|cc|c}
\hline

\multicolumn{1}{c|}{} & IND intents $\mathbf{\uparrow}$  & OOD intents $\mathbf{\uparrow}$ & ALL intents $\mathbf{\uparrow}$   \\ \hline
E2E                   & 5.25	       & 2.55	        &  3.81    \\ 
DPL(ours)           & \textbf{5.78}        & \textbf{2.71}        & \textbf{4.09} \\ \hline
\end{tabular}%
}
\caption{Cluster compactness of different methods.}
\label{tab:distance}
\end{table}

\begin{figure*}[t]
    \centering
    \resizebox{0.98\linewidth}{!}{
    \includegraphics{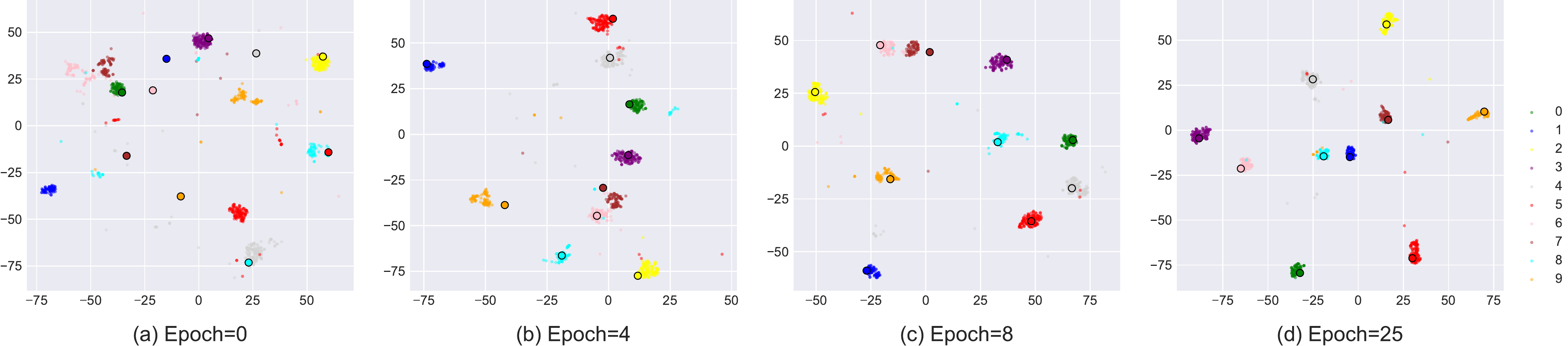}}
    \caption{intent and prototypes visualization of different training epochs for our proposed DPL method.}
    \label{prototypes_process}
\end{figure*}

\begin{table}[t]
\centering
\resizebox{0.45\textwidth}{!}{%
\begin{tabular}{l|c|c|c}
\hline
      Models         & OOD ACC   & OOD F1   & ALL F1   \\ \hline
E2E &72.28	&73.28	&84.10 \\ \hline
DPL($\mathcal{L}_{PCL}+\mathcal{L}_{ins}$)       &\textbf{74.38}	&\textbf{75.46}	&\textbf{85.34} \\
\quad-w/o $\mathcal{L}_{ins}$       &73.28	&74.25	&84.51 \\
\quad-w/o $\mathcal{L}_{PCL}$       &73.97	&74.16	&84.27 \\ \hline
$\mathcal{L}_{SCL}$       &71.15	&70.47	&83.10 \\
$\mathcal{L}_{PCL}+\mathcal{L}_{SCL}$       &71.39	&72.16	&83.86 \\
 \hline
\end{tabular}%
}
\caption{Ablation study of different representation learning objective for DPL.}
\label{tab:ablation}
\end{table}

A cluster-friendly intent representation is very important for the pseudo label disambiguation of the generalized intent discovery task. PCL can get closely aligned cluster distribution for similar samples, which is beneficial for prototype-based label disambiguation. Firstly, we quantitatively compare the cluster compactness learned by DPL and E2E. We calculate the intra-class and inter-class distances following \citet{feng2021rethinking}. For the intra-class distance, we calculate the mean value of the euclidean distance between each sample and its class center. For the inter-class distance, we calculate the mean value of the euclidean distance between the center of each class and the center of other classes. We report the ratio of inter-class and intra-class distance in Table \ref{tab:distance}. The higher the value, the clearer the boundary between different intent categories. The results show that PCL learns better intent representation, which explains why the DPL method can obtain more accurate pseudo labels.
In order to more intuitively analyze the effect of PCL in representation learning, we perform intent visualization of E2E and DPL methods, as shown in Fig \ref{final_visualization}. We can see that the DPL framework adopts PCL for representation learning, which can obtain compact cluster (see "black" and "blue" points). In addition, we can observe that clusters learned by E2E method are concentrated in the upper right part, while DPL can obtain are more evenly distributed clusters.
To see the evolution of our DPL method in the training process, we show a visualization at four different timestamps in Fig \ref{prototypes_process}. We can see that samples of different intents are mixed in the representation space at the begining, and cannot form compact clusters. As the training process goes, the boundary of different intent clusters becomes clearer and the learned class prototypes gradually falls in the center of the corresponding intent cluster.

\subsection{Ablation Study}
\label{Ablation}

To understand the effect of different contrastive learning objectives on our DPL framework, we perform ablation study in Table \ref{tab:ablation}. In our DPL framework, we jointly optimized $\mathcal{L}_{PCL}$ and $\mathcal{L}_{ins}$ to achieve the best performance. Then we remove $\mathcal{L}_{PCL}$ and $\mathcal{L}_{ins}$ respectively, and find that compared with the joint optimization, the performance drop to a certain extent, but both are better than the baseline. This shows that both prototypical contrastive learning and instance-level contrastive learning can learn discriminative intent representations and facilitate pseudo label disambiguation.

In addition, we also explored the adaptability of the commonly used supervised contrastive learning (SCL) in the DPL framework. We find that the performance of $\mathcal{L}_{SCL}$ is significantly lower than that of $\mathcal{L}_{PCL}$ and $\mathcal{L}_{ins}$. We argue that this is because SCL draws similar samples closer and pushes apart dissimilar samples, but it lacks the interaction between samples and class prototypes in the training process, and there is gap with the subsequent prototype-based label disambiguation.

\subsection{Effect of Moving Average Coefficient}
\label{Hyper}

\begin{figure}[t]
    \centering
    \resizebox{.40\textwidth}{!}{
    \includegraphics{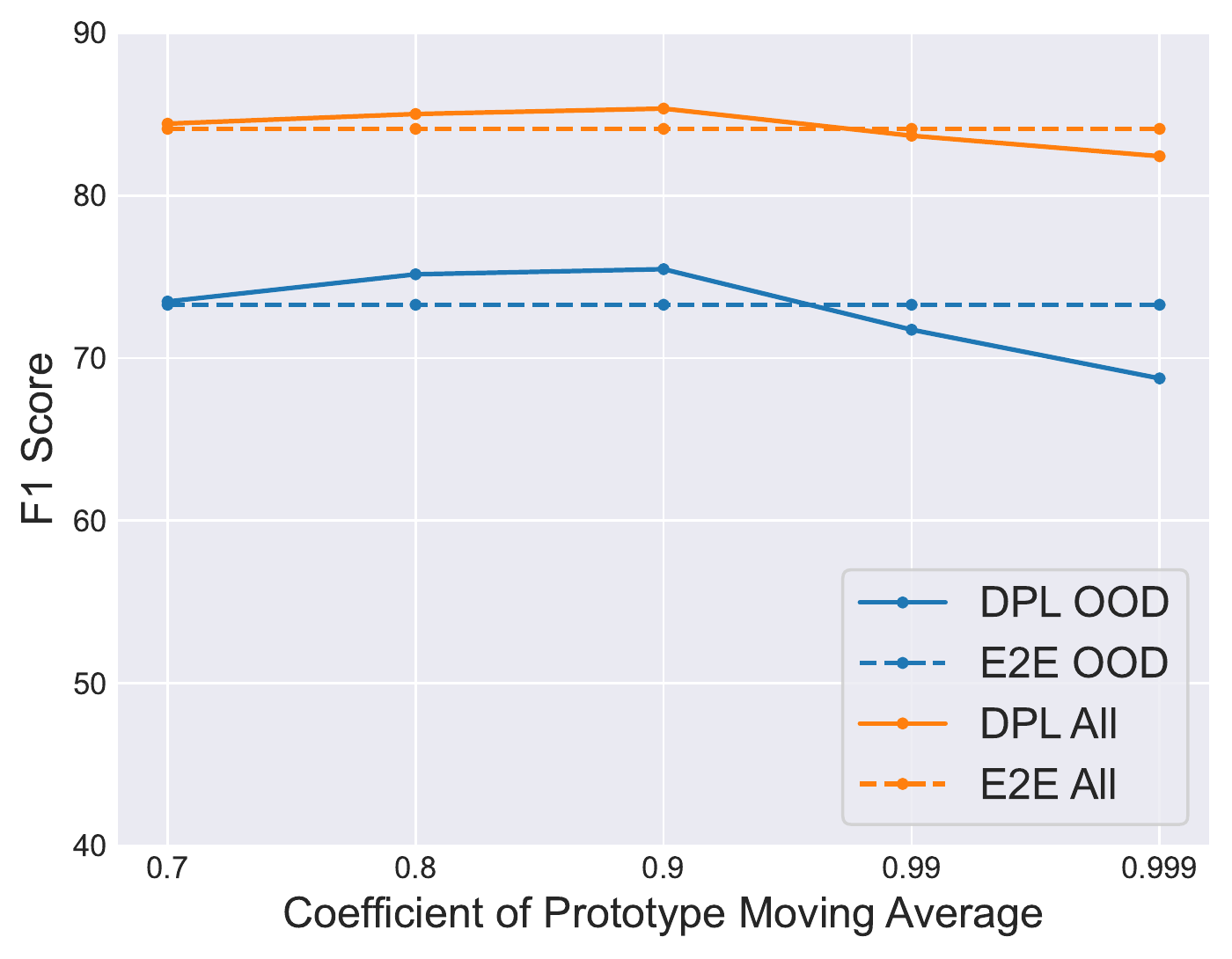}}
    \caption{Effect of the prototype moving average coefficient $\gamma$ on our DPL method. We conduct experiments on GID-SD dataset.}
     \vspace{-0.6cm}
    \label{m_proto}
\end{figure}

Fig \ref{m_proto} shows the effect of different prototype moving average coefficient $\gamma$ on our DPL method. Results show that $\gamma$ = 0.9 gets the best performance on GID-SD. Our DPL method with $\gamma$ in (0.7, 0.95) outperforms SOTA baseline, which proves DPL method is robust for different $\gamma$.
In addition, we observe that when $\gamma$ is greater than 0.9, the performance of DPL decreases significantly. We argue that this is because a large $\gamma$ will slow the speed of prototypes moving to the center of the corresponding clusters, resulting in getting poor prototypes, which hinders pseudo label disambiguation.

\subsection{Estimate the Number of OOD intents}
\label{auto-k}

In standard GID setting, we assume that the number of OOD classes is ground-truth. However, in the real applications, the number of OOD clusters often needs to be estimated automatically. We use the same OOD cluster number estimation strategy as \citet{Zhang2021DiscoveringNI, mou-etal-2022-generalized}. The results are showed in Table \ref{tab:k}. It can be seen that when the number of OOD clusters is inaccurate, all methods have a certain decline, but our DPL method still significantly outperforms all baselines, and even the improvement is more obvious, which also proves that DPL is more robust and practical.


\begin{table}[]
\centering
\resizebox{0.45\textwidth}{!}{%
\begin{tabular}{l|c|c|c|c}
\hline
\multicolumn{1}{c|}{} & OOD ACC     & OOD F1     & ALL F1    & K  \\ \hline
DeepAligned             &69.11	&69.72	&82.10  & 31 \\ 
End-to-End       &72.28	&73.28	&84.10  & 31   \\
DPL(ours) &\textbf{74.38}	&\textbf{75.46}	&\textbf{85.34}  & 31 \\ \hline
DeepAligned             &62.50	&59.74	&77.39  & 26 \\ 
End-to-End       &66.29	&61.55	&78.57  & 26   \\
DPL(ours) &\textbf{70.81}	&\textbf{67.57}	&\textbf{81.17}  & 26 \\ \hline
\end{tabular}%
}
\caption{Estimate the number of OOD classes. We take GID-SD as an example. K=26 is the estimated number compared to ground-truth number 31.}
\label{tab:k}
\end{table}

 \subsection{Effect of different OOD ratios}
\label{OOD ratio}
In Fig \ref{fig:ood_ratio}, we compare the effect of different OOD ratios on various methods.
The larger the OOD ratio means the fewer the IND categories and the more the OOD categories. On the one hand, it reduces the available prior knowledge of IND intents, and on the other hand, it is more difficult to distinguish the unlabeled OOD intents. The experimental results show that the performance of all methods decrease significantly  as the OOD ratio increases. However, we find that when the OOD ratio is large, our DPL method has a more obvious improvement compared with other baselines, which shows that our method is more robust to different OOD ratios, and DPL decouples pseudo label disambiguation and representation learning, which can more effectively use the prior knowledge of IND intents and learn the discriminative intent representations, which improves the reliability of pseudo labels. 

\begin{figure}[t]
    \centering
    \resizebox{0.95\linewidth}{!}{
    \includegraphics{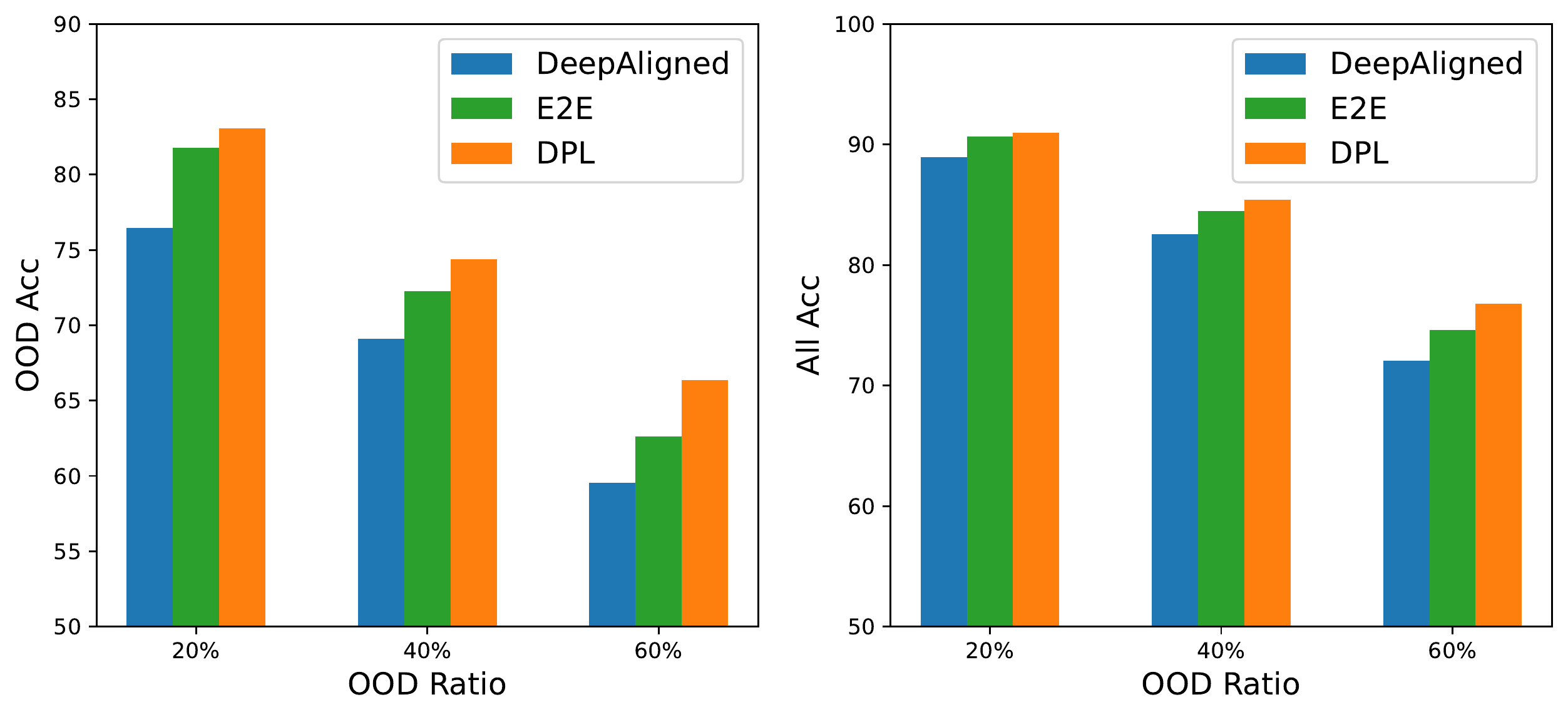}}
    \caption{The effect of different OOD ratios on the performance of each GID method.}
    \label{fig:ood_ratio}
\end{figure}

\subsection{Effect of imbalanced OOD data}
As mentioned in the previous work \cite{mou-etal-2022-generalized}, E2E introduces a method based on optimal transport \cite{Cuturi2013SinkhornDL, caron2020unsupervised, Fini2021AUO} to calibrate the output logits of the joint classifier before swapped prediction. This assumes that the unlabeled OOD samples in each batch are evenly distributed to $M$ OOD categories. However, it is hard to ensure that the unlabeled OOD data in the real scene is class-balanced, and sometimes even the long-tailed distribution. The experimental results in Fig \ref{fig_imbalanced} show that the E2E method has a significant performance degradation in the case of OOD class-imbalanced. In contrast, our proposed DPL framework adopts a prototype-based label disambiguation method, and it doesn't rely on the assumption of class-distribution assumption. Therefore, it is significantly better than E2E in the OOD class-imbalanced scenario.

However, we also observed that when the imbalance factor increased, the performance of our DPL method decline more significantly compared with DeepAligned. We think that this is because DPL method needs to use unlabeled OOD data to learn the discriminative representations and class prototypes. When the imbalance factor increases, the number of samples for OOD intent categories will become smaller, which is not conducive to learning the cluster-friendly representations and class prototypes. We can alleviate this problem by increasing the number of samples through data augmentation. We will leave this to future work.

\section{Related Work}

\label{imbalanced}
\begin{figure}[t]
    \centering
    \resizebox{1.0\linewidth}{!}{
    \includegraphics{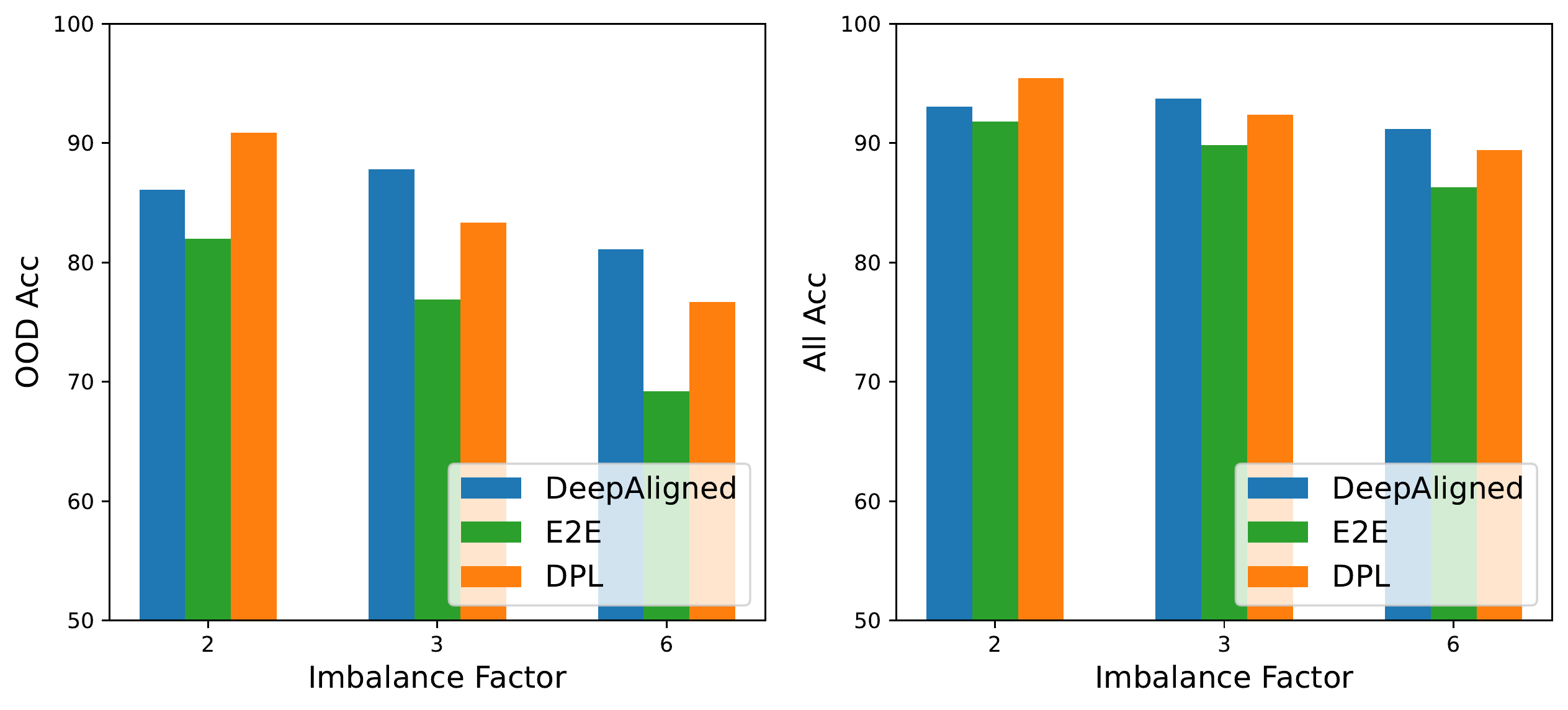}}
    \caption{The effect of different imbalance factors of
OOD data on the performance of each GID method.}
    \label{fig_imbalanced}
\end{figure}

\textbf{Generalized Intent Discovery} Existing intent classification models have little to offer in an open-world setting, in which many new intent categories are not pre-defined and no labeled data is available. These models can only recognize limited in-domain (IND) intent categories. \citet{lin-xu-2019-deep, xu-etal-2020-deep} propose the OOD intent detection task to identify whether a user query falls outside the range of a pre-defined intent set. Further, OOD intent discovery task (also known as new intent discovery) \cite{Lin2020DiscoveringNI,Zhang2021DiscoveringNI} is proposed to cluster unlabeled OOD data.
\citet{mou-etal-2022-generalized} proposes the Generalized Intent Discovery (GID) task, which aims to simultaneously classify a set of labeled IND intents while discovering and recognizing new unlabeled OOD types incrementally. 

\textbf{Prototype-based Learning} Prototype-based metric learning methods have been promising approaches in many applications.  \citet{Snell2017PrototypicalNF} first proposes Prototypical Networks (ProtoNet) which introduces prototypes into deep learning. Specifically, ProtoNet calculates prototype vectors by taking the average of instance vectors and makes predictions by metric-based comparisons between prototypes and query instances. \citet{Li2020PrototypicalCL} proposes 
self-supervised prototype representation learning by using prototypes as latent variables. Learning good representations also helps weakly supervised learning tasks, including noisy label learning \cite{Li2020MoProWS}, semi-supervised learning \cite{Zhang2021SemisupervisedCL}, partial label learning \cite{Wang2022PiCOCL}, etc. Inspired by these methods, we propose a decoupled prototype learning framework (DPL) to decouple pseudo label disambiguation and representation learning for GID.

\section{Conclusion}
In this paper, we propose a decoupled prototype learning (DPL) framework for generalized intent discovery. We introduce prototypical contrastive representation learning and prototype-based label disambiguation method to decouple representation learning and pseudo label disambiguation. Theoretical analysis and extensive experiments prove that our method can learn discriminative intent representations and prototypes, which facilitates pseudo label disambiguation. We will explore broader applications of DPL method in the future.

\section*{Limitations}

This paper mainly focuses on the generalized intent discovery (GID) task in task-oriented dialogue systems. Our proposed Decoupled Prototype Learning (DPL) framework 
well decouple pseudo label disambiguation and representation learning through protopical contrastive learning and prototype-based label disambiguation, and achieves SOTA performance on three GID benchmark datasets. However, our work also have several limitations: (1) We only verified the effectiveness of our DPL framework on GID task, but the adaptability of DPL in more unsupervised / semi-supervised settings, such as unsupervised clustering and OOD intent discovery, is worth further exploration.
(2) We follow standard experiment settings as previous work, and assume that each OOD sample must belong to a corresponding intent cluster. However, a more realistic scenario is that there may be noise samples in the OOD data. These noise samples do not actually belong to any cluster/category and are some outliers. We leave the noise OOD issue to the future work.
(3) Our experiments in Appendix \ref{imbalanced} find that the performance of the DPL method decreases significantly when the imbalance factor of unlabeled OOD data increases. How to improve the performance of GID model on the long tailed unlabeled data is also a problem worthy of attention in the future.

\section*{Acknowledgements}
We thank all anonymous reviewers for their helpful comments and suggestions. This work was partially supported by National Key R\&D Program of China No. 2019YFF0303300 and Subject II No. 2019YFF0303302, DOCOMO Beijing Communications Laboratories Co., Ltd, MoE-CMCC "Artifical Intelligence" Project No. MCM20190701.


\bibliography{anthology,custom}
\bibliographystyle{acl_natbib}

\appendix

\section{Datasets}
\label{appendix_data}
Table \ref{tab:dataset1} shows the statistics of the original datasets Banking and CLINC, where each class in CLINC has the same number of samples but Banking is class-imbalanced. For the three GID datasets GID-SD, GID-CD and GID-MD, We show the detailed statistics in Table \ref{tab:dataset2}. Due to Banking is class-imbalanced and we conducted three random partitions, we report the average of the sample number of GID-SD.

\begin{table*}[t]
	\centering
 \resizebox{0.9\textwidth}{!}{%
	\begin{tabular}{ccccccc}
		\toprule
		Dataset & Classes &Training & Validation & Test & Vocabulary & Length (max / mean)\\
		\midrule
		Banking & 77 & 9003 & 1000 & 3080 & 5028&79/11.91\\
		CLINC & 150 & 18000 & 2250 &2250 &7283&28/8.31\\
		\bottomrule
	\end{tabular}%
 }
 	
 \caption{Statistics of Banking and CLINC datasets.}
 \label{tab:dataset1}
\end{table*}

\begin{table*}[t]
	\centering
\resizebox{0.95\textwidth}{!}{%
	\begin{tabular}{cccccc}
		\toprule
		Dataset &IND/OOD Classes & IND/OOD Domains & IND/OOD Training & IND/OOD Validation&IND/OOD Test\\
        \midrule
		GID-SD & 46/31 & 1/1 & 5346/3657 &593/407 & 1840/1240\\
        GID-CD & 90/60 & 6/4 & 7200/4800 &1350/900 & 1350/900\\
		GID-MD & 90/60 & 10/10 & 7200/4800 &1350/900 & 1350/900\\	
		\bottomrule
	\end{tabular}%
 }
 \caption{Statistics of GID-SD, GID-CD and GID-MD datasets.}
 \label{tab:dataset2}
\end{table*}

\section{Baselines}
\label{appendix_baseline}
The details of baselines are as follows:

\textbf{k-means} is a pipeline method which first uses kmeans \cite{MacQueen1967SomeMF} to cluster OOD data and obtains pseudo OOD labels, and then trains a new classifier together with IND data.

\textbf{DeepAligned} is similar to k-means, the difference is that the clustering algorithm adopts DeepAligned \cite{Zhang2021DiscoveringNI}, which uses an alignment strategy to tackle the label inconsistency problem during clustering assignments.

\textbf{DeepAligned-Mix} \cite{mou-etal-2022-generalized} is an extended method from DeepAligned for GID task. In each iteration, it firstly mix up IND and OOD data together for clustering  using k-means and an alignment strategy and then uses a unified cross-entropy loss to  optimize the model. In the inference stage, instead of using k-means for clustering, DeepAligned-Mix use the classification head of the new classifier to make predictions.

\textbf{E2E} \cite{mou-etal-2022-generalized} mixes IND and OOD data in the training process and simultaneously learns pseudo OOD cluster as signments and classifies all classes via self-labeling. Given an input query, E2E connects the encoder output through two independent projection layers, IND head and OOD head, as the final logit and  optimize the model through the unified classification loss, where the OOD pseudo label is obtained through swapped prediction\cite{caron2020unsupervised}.

\section{Details of derivation process}
\label{appendix_theory}

\subsection{Derivation process of equation \ref{e-step}}

In the GID task, likelihood function is hard to be directly optimized, so we need to introduce a probability density function ${q}_{i}(j)$ to represent the probability that sample ${x}_{i}$ belongs to intent $j$. The detailed derivation process is as follows:

\begin{equation}
\begin{aligned}
\theta^* & =\underset{\theta}{\operatorname{argmax}} \sum_{i=1}^{n+m} \log P\left(x_i \mid \theta\right) \\
& =\underset{\theta}{\operatorname{argmax}} \sum_{i=1}^{n+m} \log \sum_{j \in y_{\text {all }}} P\left(x_i, j \mid \theta\right) \\
& =\arg \max \sum_{i=1}^{n+m} \log \sum_{j \in y_{a l l}} q_i(j) \frac{P\left(x_i, j \mid \theta\right)}{q_i(j)} \\
& \geqslant \underset{\theta}{\arg \max } \sum_{i=1}^{n+m} \sum_{j \in y_{a l l}} q_i(j) \log \frac{P\left(x_i, j \mid \theta\right)} {q_{i}(j)} \\
\end{aligned}
\end{equation}

\subsection{Derivation process of equation \ref{m-step}}
\begin{equation}
\begin{aligned}
& L(\theta) =\max \sum_{i=1}^{n+m} \sum_{j \in y_{a l l}} q_i(j) \log \frac{P\left(x_i, j \mid \theta\right)}{q_i(j)} \\
& =\max \sum_{i=1}^{n+m} \sum_{j \in y_{a l l}} q_i(j) \log P\left(x_i, j \mid \theta\right) \\
& \approx \max \sum_{i=1}^{n+m} \sum_{j \in y_{a l l}} q_i(j) \log P\left(x_i \mid j, \theta\right) \\
& =\max \sum_{i, j} q_i(j) \log \frac{\exp \left(\frac{-\left(z_i-\mu_j\right)^2}{2 \sigma_j^2}\right)}{\sum_{r \in y_{a l l}} \exp \left(\frac{-\left(z_i-\mu_r\right)^2}{2 \sigma_{r}^2}\right)} \\
& \approx \max \sum_{i=1}^{n+m} \sum_{j \in y_{a l l}} q_i(j) \log \frac{\exp \left(\frac{2 z_i \cdot \mu_j}{2 \sigma_j^2}\right)}{\sum_{r \in y_{a l l}} \exp \left(\frac{2 z_i \cdot \mu_r}{2 \sigma_r^2}\right)} \\
& \approx  \max \sum_{i=1}^{n+m} \sum_{j \in y_{a l l}} q_{i(j)} \log \frac{\exp \left(\frac{\boldsymbol{z_i} \cdot \boldsymbol{\mu_{j}}}{\sigma_j^2}\right)}{\sum_{r \in y_{a l l}} \exp \left(\frac{\boldsymbol{z_i} \cdot \boldsymbol{\mu_{r}}}{\sigma_r^2}\right)} \\
& \Leftrightarrow \min \mathcal{L}_{P C L} \\
\end{aligned}
\end{equation}

\begin{algorithm*}
\caption{: Decoupled Prototype Learning}
\label{algorithm_DPL}
\begin{algorithmic}[1]
\Require {training dataset $\textbf{D}^{IND}=\left\{\left({x}_{i}^{IND}, {y}_{i}^{IND}\right)\right\}_{i=1}^n$ and  $\textbf{D}^{OOD}=\left\{\left({x}_{i}^{OOD}\right)\right\}_{i=1}^m$, IND label set $\mathcal{Y}^{IND}=\{1,2, \ldots, N\}$, ground-truth number of OOD intents $M$, training epoch $E$, batch size $B$}
\Ensure  a new intent classification model, which can classify an input query to the total label set $\mathcal{Y}=\left\{1, \ldots, N, N+1, \ldots, N+M\right\}$.
\State randomly initialize the $L_{2}$-normalized prototype embedding $\mu_{j}, j = 1,2,...,N+M$.
\For {epoch = 1 to $E$}
\State {mix $\textbf{D}^{IND}$ and $\textbf{D}^{OOD}$ to get $\textbf{D}^{ALL}$}
\For {$iter$ = 0, 1, 2, ...}
\State{sample a mini-batch $\textbf{B}$ from $\textbf{D}^{ALL}$}
\State {get the $L_{2}$-normalized embedding $z_{i}$ of sample $x_{i}$ through the projection layer}
\State {align sample $x_{i}$ with prototypes $\mu_{j}$ by equation \ref{alignment}}
\State {compute $\mathcal{L}_{P C L}$ and $\mathcal{L}_{ins}$}
\Comment{\textbf{prototypical contrastive representation learning}}
\State {estimate the pseudo label $\boldsymbol{y}_{i}$ by equation \ref{disambiguation_1} and \ref{disambiguation_2}}
\Comment{\textbf{pseudo label disambiguation}}
\State {compute $\mathcal{L}_{CE}$ on the joint classifier}
\State {add $\mathcal{L}_{P C L}$, $\mathcal{L}_{ins}$ and $\mathcal{L}_{CE}$ together, and jointly optimize them}
\State{update the prototype vector by equation \ref{update}}
\EndFor
\EndFor
\end{algorithmic}
\end{algorithm*}

\section{Algorithm}
\label{appendix_algorithm}

We summarize the pseudo-code of our DPL method in Algorithm \ref{algorithm_DPL}.

\end{document}